\documentclass[a4paper,fleqn,numbers]{cas-dc}

\usepackage[]{natbib}
\usepackage{afterpage}
\usepackage{amsmath}
\usepackage{algorithm}
\usepackage{algpseudocode}
\usepackage{graphicx}
\usepackage{textcomp}
\usepackage{xcolor}
\usepackage{pifont}
\usepackage{booktabs} 
\usepackage{multirow}
\usepackage[normalem]{ulem}
\useunder{\uline}{\ul}{}
\usepackage{graphicx}
\newcommand{\OurModel}[1]{MAT-Adv}
\newcommand{\firstpara}[1]{\noindent\textbf{{#1}}~~}
\newcommand{\firstparaNo}[1]{\noindent\textbf{{#1}}}

\begin{document}
\let\WriteBookmarks\relax
\def\floatpagepagefraction{1}
\def\textpagefraction{.001}

\title[mode = title]{Transferable and Undefendable Point Cloud Attacks via
Medial Axis Transform}


\author[1]{Keke Tang}\corref{cor1}
\author[1]{Yuze Gao}
\author[2]{Weilong Peng}
\author[3]{Xiaofei Wang}
\author[2]{Meie Fang}
\author[4]{Peican Zhu}

\address[1]{Cyberspace Institute of Advanced Technology, Guangzhou University, Guangzhou, Guangdong 510006, China}
\address[2]{School of Computer Science and Cyber Engineering, Guangzhou University, Guangzhou, Guangdong 510006, China}
\address[3]{Department of Automation, University of Science and Technology of China,
Hefei, Anhui 230052, China}
\address[4]{School of Artificial Intelligence, Optics and
Electronics (iOPEN), Northwestern
Polytechnical University,
Xi’an, Shaanxi 710072, China}

\cortext[cor1]{Corresponding author.}

\begin{abstract}[S U M M A R Y]
Studying adversarial attacks on point clouds is essential for evaluating and improving the robustness of 3D deep learning models. However, most existing attack methods are developed under ideal white-box settings and often suffer from limited transferability to unseen models and insufficient robustness against common defense mechanisms.
In this paper, we propose MAT-Adv, a novel adversarial attack framework that enhances both transferability and undefendability by explicitly perturbing the medial axis transform (MAT) representations, in order to induce inherent adversarialness in the resulting point clouds.
Specifically, we employ an autoencoder to project input point clouds into compact MAT representations that capture the intrinsic geometric structure of point clouds. By perturbing these intrinsic representations, MAT-Adv introduces structural-level adversarial characteristics that remain effective across diverse models and defense strategies.
To mitigate overfitting and prevent perturbation collapse, we incorporate a dropout strategy into the optimization of MAT perturbations, further improving transferability and undefendability.
Extensive experiments demonstrate that MAT-Adv significantly outperforms existing state-of-the-art methods in both transferability and undefendability. Codes will be made public upon paper acceptance.
\end{abstract}


	
		

\maketitle

\section{Introduction}\label{sec:intro}
The rapid advancement of deep learning has positioned deep neural networks (DNNs) as the foundation for 3D point cloud analysis~\cite{guo2020deep}. Despite their impressive performance, these models have been shown to be highly vulnerable to adversarial attacks, where subtle and often imperceptible perturbations to the input point clouds can lead to severe misclassifications~\cite{Xiang-2019-Generating,liu-2019-extending,liu-2022-imperceptible(ITA),Tang-2024-FLAT}. This vulnerability raises serious concerns in safety-critical applications such as autonomous driving and service robotics, where model reliability is paramount. Consequently, investigating adversarial attacks is crucial for understanding the vulnerabilities of DNN-based point cloud classifiers, assessing their security risks, and ultimately facilitating future robustness enhancements.


Due to the inherently unstructured nature of point clouds, performing adversarial attacks on them is particularly challenging, as small modifications can easily result in perceptible distortions. 
To address this issue, existing methods commonly incorporate geometric constraints such as the \(l_2\) norm, Chamfer distance, or Hausdorff distance~\cite{wen-2020-geometry(geo)}, to maintain geometric fidelity.  
In addition, several works introduce structural priors, including symmetry~\cite{tang2024symattack} and uniformity~\cite{Tang-2024-FLAT}, to guide perturbations in a perceptually coherent manner.  
Other approaches constrain perturbations along specific geometric directions, such as surface normals~\cite{liu-2022-imperceptible(ITA)} or tangent planes~\cite{Huang-2022-shape(SI-Adv)}, in order to better align with the local shape manifold.
While these techniques often perform well in white-box settings where model gradients are accessible, real-world scenarios are typically black-box, in which model architectures and parameters are unknown.  
In such cases, the transferability of adversarial examples—their ability to mislead unseen models—becomes particularly critical.  
Furthermore, as defense mechanisms such as denoising are commonly deployed in practice, adversarial examples must also demonstrate strong undefendability, that is, robustness against such countermeasures.  
Therefore, simultaneously achieving high transferability and strong undefendability is essential for ensuring the practical effectiveness of adversarial attacks.


 \begin{figure}[!t] 
\centering
\includegraphics[width=\linewidth]{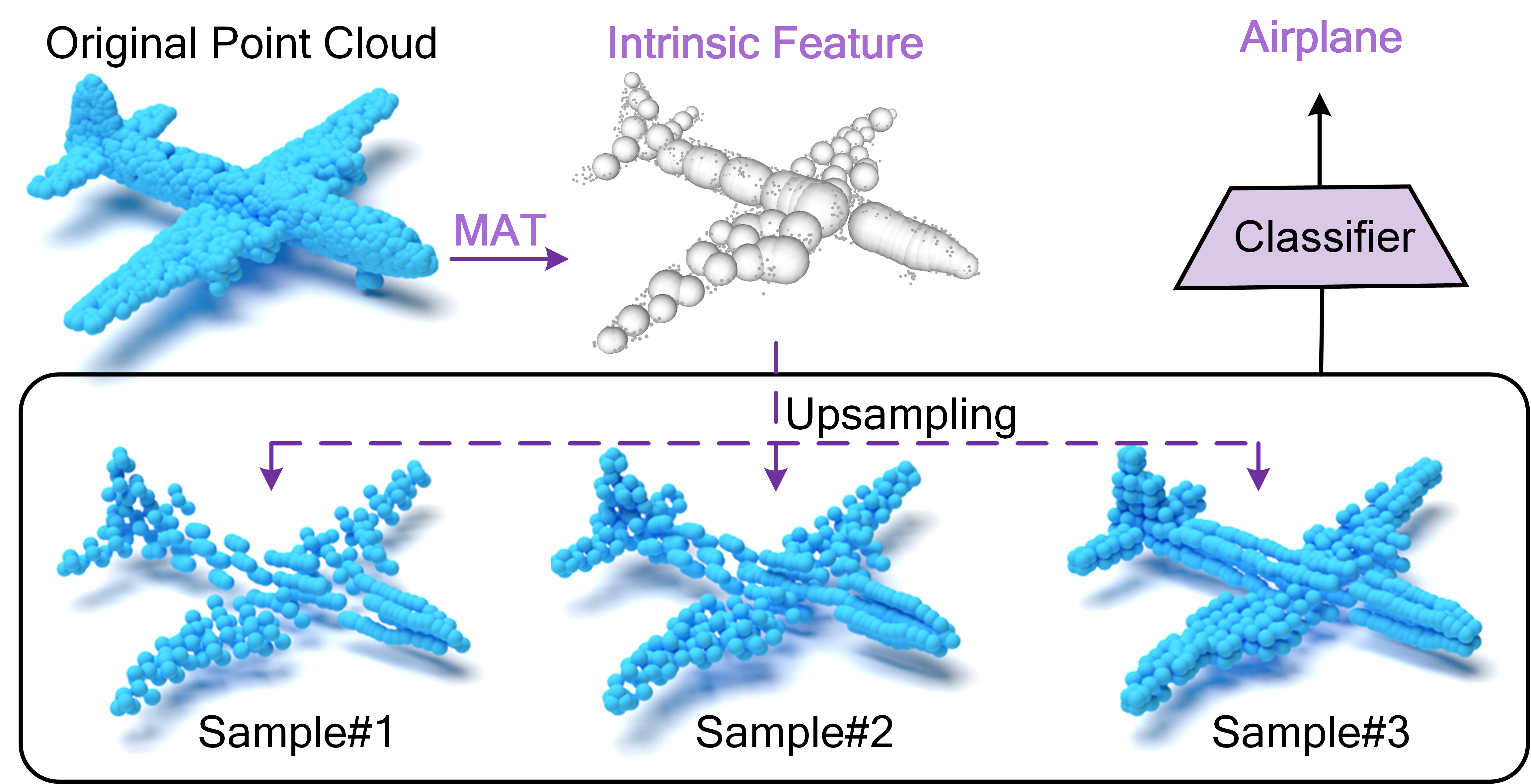}
\vspace{-1mm}
 \caption{
Given a point cloud, we extract its medial axis transform (MAT) representation. The three upsampled point clouds generated from the medial spheres retain the  characteristics of the original cloud, and can be correctly classified by a classifier trained on the original data, e.g., PointNet, validating that the MAT representation captures intrinsic features of the point cloud.
}
\label{fig:teaser} 
\end{figure}


Simultaneously achieving transferability and undefendability remains a fundamental challenge, as it requires adversarial point clouds to exhibit \textit{inherent adversarialness}.
AdvPC~\cite{hamdi2020advpc} addresses this challenge by incorporating a denoising autoencoder as a constraint, requiring the reconstructed output to remain adversarial when generating perturbations. This strategy encourages perturbations to survive structural reconstruction, thereby explicitly promoting inherent adversarialness. Building on the observation that point clouds are essentially sampled from 2-manifold surfaces embedded in 3D Euclidean space, Mani-Adv~\cite{Tang-2023-ManifoldAttack} maps point clouds to a  two-dimensional parameter domain and applies perturbations in that space. This approach better aligns with the intrinsic structure of the underlying surface and improves both transferability and undefendability. However, although it operates in a space that reflects geometric properties, it does not explicitly model or control the underlying geometric structures. To further enhance inherent adversarialness, we aim to construct a more interpretable and controllable intrinsic representation of point cloud geometry. One promising candidate is the medial axis transform (MAT)~\cite{amenta2001power}, which represents a shape as the union of its maximal inscribed spheres. As shown in Fig.~\ref{fig:teaser}, point clouds reconstructed from MAT representations consistently preserve essential shape characteristics, demonstrating the structural fidelity and expressive power of this representation. Motivated by this observation, we propose to directly perturb the MAT representation in order to induce stronger inherent adversarialness, thereby generating adversarial point clouds with improved transferability and undefendability.

In this paper, we propose a novel framework for generating transferable and undefendable adversarial attacks on point clouds, termed \OurModel{}. The core idea is to explicitly leverage the intrinsic properties encoded by the medial axis transform (MAT). Specifically, we design an autoencoder that maps input point clouds into a compact MAT-based representation that captures essential geometric structures. Adversarial perturbations are applied to the MAT representations, and the perturbed representations are decoded into adversarial point clouds by sampling points on the medial spheres. This process effectively embeds intrinsic adversarial characteristics into the generated shapes, thereby enhancing both transferability and undefendability. To mitigate overfitting to specific models and prevent perturbation collapse, we introduce a dropout strategy that randomly masks perturbations
on the MAT representations
during adversarial optimization. This mechanism strengthens the robustness of individual perturbations and leads to improved transferability and resistance to  defenses. We validate the effectiveness of \OurModel{} on several standard 3D point cloud datasets and across a diverse set of commonly used DNN classifiers, under a variety of defense mechanisms. Extensive experimental results demonstrate that our approach significantly outperforms existing state-of-the-art methods in both transferability and undefendability.

Overall, our contributions are summarized as follows:

\begin{itemize} 

\item We identify that achieving strong transferability and undefendability requires adversarial point clouds to exhibit inherent adversarialness, which can be achieved by  perturbing intrinsic representations of point cloud geometry.

\item We propose MAT-Adv, a novel autoencoder-based attack framework that encodes input point clouds into MAT representations and applies perturbations directly to these representations to generate adversarial examples.

\item We demonstrate through extensive experiments that our proposed \OurModel{} achieves superior transferability and undefendability compared to existing methods.

\end{itemize}

\section{Related Work}

\subsection{Adversarial Attacks on 3D Point Clouds}

Adversarial attacks~\cite{Yang-2019-AdvAttackAndDefense,wei2024physical}, initially developed for 2D image classification, have been extended to 3D point clouds, introducing unique challenges due to the unordered and irregular nature of 3D data. Existing attack methods are typically categorized into three types: addition-based, which insert new points to induce misclassification~\cite{Xiang-2019-Generating}; deletion-based, which remove critical points to disrupt model predictions~\cite{Zheng-2019-PCDSaliency,Wicker-2019-IterSaliencyOcc}; and perturbation-based, which apply small coordinate modifications to mislead the model~\cite{zhao2020isometry,Tang-2024-FLAT,tang2024manifold,lou2024hide}. Among these, perturbation-based attacks have received the most attention due to their simplicity and strong effectiveness.

Early works such as~\cite{Xiang-2019-Generating,liu-2019-extending} adapted classical 2D techniques like FGSM~\cite{Goodfellow-2014-FGSM} and C\&W~\cite{Carlini-2017-cw} to the 3D setting. To improve imperceptibility, subsequent studies introduced various constraints. Some methods incorporate geometric priors such as curvature~\cite{wen-2020-geometry(geo)}, symmetry~\cite{tang2024symattack}, and uniformity~\cite{Tang-2024-FLAT}, while others constrain perturbations along specific geometric directions, such as surface normals~\cite{liu-2022-imperceptible(ITA)} or tangent planes~\cite{Huang-2022-shape(SI-Adv)}.
While existing research primarily focuses on white-box settings and perceptual stealthiness, our work explores a complementary and underexplored dimension: the transferability and undefendability of adversarial examples. 

\subsection{Transferable and Undefendable Point Cloud Attacks}

Several works have studied the transferability of adversarial point cloud attacks. Liu et al.~\cite{liu-2022-imperceptible(ITA)} proposed an adversarial transformation model that learns to generate harmful distortions and then trains adversarial examples to resist these distortions, thereby improving their transferability to black-box models.
Liu et al.~\cite{liu2022boosting} introduced a frequency-domain attack based on graph Fourier transforms, where perturbing low-frequency components was shown to enhance transferability.
He et al.~\cite{he2023generating} proposed decomposing adversarial perturbations into primary and sub-perturbations, which are jointly optimized to improve transferability.
Chen et al.~\cite{ANF} developed ANF, a noise factorization-based method that simultaneously optimizes adversarial noise along with its positive and negative components in the feature space. By relying only on partial network parameters, ANF reduces dependence on the surrogate model and improves transferability.

A smaller number of methods consider both transferability and undefendability. Hamdi et al.~\cite{hamdi2020advpc} proposed AdvPC, which leverages a denoising autoencoder during iterative optimization. By enforcing adversariality on both the original and reconstructed point clouds, AdvPC reduces overfitting to the surrogate model and improves robustness against defenses.
Mani-Adv~\cite{Tang-2023-ManifoldAttack} maps point clouds to a continuous 2D parameter space based on the assumption that point clouds are sampled from 2-manifold surfaces embedded in 3D Euclidean space. Perturbing this parameter space better aligns with intrinsic geometry, enhancing  transferability and undefendability.

In this work, we  advance this line of research by introducing an adversarial attack framework that operates on the medial axis transform (MAT), an intrinsic representation of point cloud geometry. Our approach achieves improved transferability and undefendability by leveraging this compact and interpretable structure.

\subsection{DNN-based 3D Point Cloud Classification}

Recent advances in deep learning have significantly advanced 3D point cloud classification. Early methods relied on voxelization and 3D CNNs, which were limited by high computational cost and resolution requirements~\cite{Maturana-2015-voxnet}. The introduction of PointNet~\cite{Qi-2017-Pointnet} and its hierarchical extension PointNet++~\cite{Qi-2017-Pointnet++} enabled direct processing of raw point clouds, leading to a wide range of architectures, including point-wise convolutional networks~\cite{Wu-2019-Pointconv,Li-2018-PointCNN}, graph-based models~\cite{Wang-2019-DGCNN,Zhao-2019-Pointweb}, and more recently, Transformer- and Mamba-based designs~\cite{zhao2021PT,wu2024pT3,liang2024pointmamba}, which have collectively improved classification performance.
In this work, we develop transferable and undefendable adversarial attacks specifically targeting these point cloud classification models.

\subsection{Medial Axis Transform}

The medial axis transform (MAT)~\cite{amenta2001power, sun2015medial} is a classical tool in 3D shape analysis, known for its ability to capture the intrinsic structural properties of geometric objects. It has been widely used in computer graphics and robotics. In graphics, MAT facilitates shape simplification~\cite{tam2003shape}, feature extraction~\cite{lerner1995medial}, and mesh generation~\cite{sun2015medial} by providing skeletal representations of complex geometry. In robotics, it supports collision detection and path planning by compactly encoding free space and object boundaries~\cite{gayle2005path}.
In this work, we leverage MAT to construct a compact and interpretable intrinsic representation of point cloud geometry, which serves as the basis for designing transferable and undefendable adversarial attacks.

\begin{figure*}[!t] 
\centering
\includegraphics[width=0.82\linewidth]{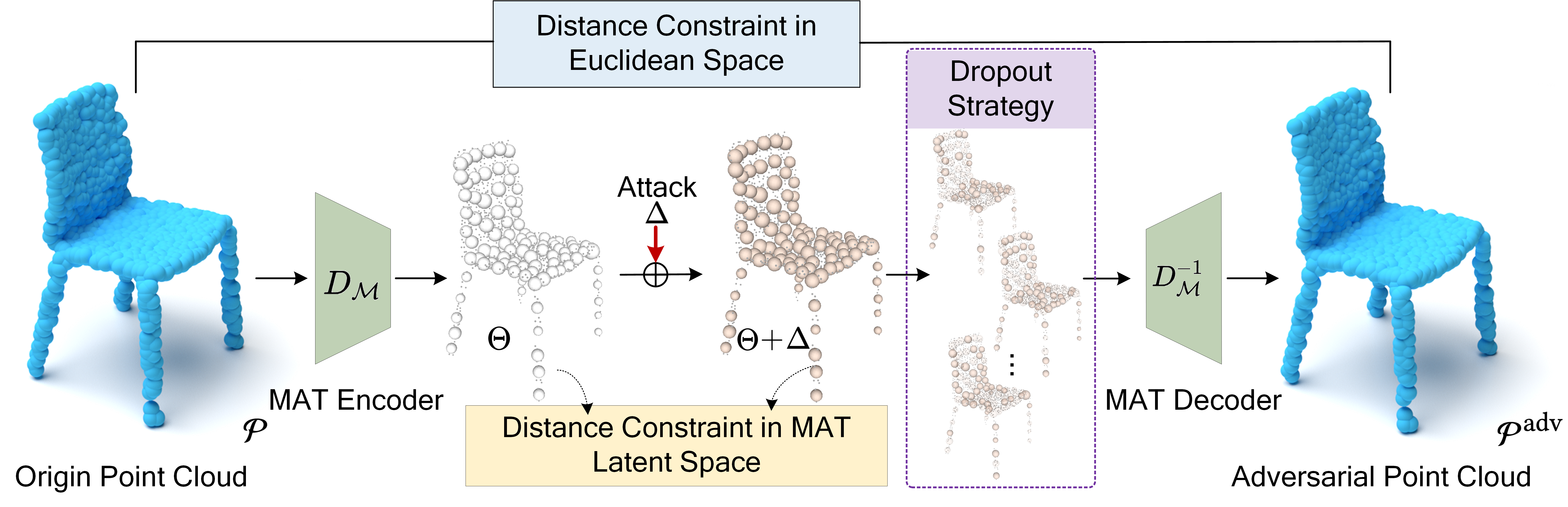}
\caption{
Illustration of our transferable and undefendable point cloud attack framework via medial axis transform (\OurModel{}).   Given an input point cloud \(\mathcal{P}\), we extract its MAT representation using an encoder \(D_{\mathcal M}\), apply targeted perturbations with a dropout strategy to regularize the optimization, and decode the result via \(D_{{\mathcal M}^{-1}}\) to obtain the adversarial point cloud \(\mathcal{P}^{\text{adv}}\).
}
\label{fig:method} 
\end{figure*}


\section{Problem Formulation}

\firstpara{Typical  Adversarial Attacks.}
Given a point cloud \(\mathcal{P} \in \mathbb{R}^{N \times 3}\) and its associated class label \(y \in \{1, \ldots, Z\}\), an adversarial attack aims to mislead a 3D DNN classifier \(f_s\) by generating an adversarial point cloud \(\mathcal{P}^{\text{adv}}\). Formally, this can be expressed as:
\begin{equation}
\setlength\abovedisplayskip{3pt}
\setlength\belowdisplayskip{3pt}
\min_{\delta} L_{\text{mis}}(f_s, \mathcal{P} + \delta, y) + \lambda_1 D(\mathcal{P}, \mathcal{P} + \delta),
\end{equation}
where \(L_{\text{mis}}(\cdot,\cdot,\cdot)\) denotes the loss promoting misclassification (e.g., negative cross-entropy), \(D(\cdot,\cdot)\) enforces imperceptibility constraints on perturbations, and \(\lambda_1\) balances these two terms. While our primary focus is on untargeted attacks, the formulation can be readily extended to targeted settings.

\firstpara{Adversarial Attacks in Practical Scenarios.}  
Most existing adversarial attacks are designed under a white-box setting, where the victim classifier \(f_s\) is known in advance and fully accessible.  
In contrast, practical scenarios are significantly different: the true target classifier \(f_t\) is typically unknown, and adversarial examples must often be generated using a surrogate model \(f_s\) that is accessible.  
Moreover, real-world systems usually deploy defense mechanisms—such as denoising or reconstruction-based preprocessing—denoted as \(\operatorname{Defend}(\cdot)\), to mitigate potential attacks.  
As such, a practical adversarial attack should possess both strong \textit{transferability}, enabling it to fool unseen classifiers, and high \textit{undefendability}, allowing it to bypass commonly used defenses.  
Formally, the attack objective in such a scenario can be formulated as:
\begin{equation}
\setlength\abovedisplayskip{3pt}
\setlength\belowdisplayskip{3pt}
\min_{\delta} L_{\text{mis}}(f_t \oplus \operatorname{Defend}, \mathcal{P} + \delta, y) + \lambda_1 D(\mathcal{P}, \mathcal{P} + \delta),
\label{eq:realworld}
\end{equation}
where \(f_t \oplus \operatorname{Defend}\) represents the composition of the defense mechanism with the unknown target classifier \(f_t\).

\firstparaNo{What Makes Transferability and Undefendability?}
We argue that adversarial point clouds must possess \textit{inherent adversarialness}—a quality that enables them to remain effective across unseen models and under common defenses.
Prior works have attempted to promote this property through structural considerations. AdvPC~\cite{hamdi2020advpc} employs an autoencoder and enforces that adversarialness persists after denoising. Mani-Adv~\cite{Tang-2023-ManifoldAttack} perturbs point clouds in a learned 2D manifold space to align with their underlying geometry.

In this paper, we propose a more principled and interpretable approach by perturbing intrinsic and explicitly defined geometric structures. In particular, we leverage the medial axis transform (MAT)~\cite{amenta2001power,sun2015medial} to induce inherent adversarialness, thereby enhancing both transferability and undefendability.

\firstpara{Preliminary on Medial Axis Transform.}
The medial axis transform (MAT) \(\mathcal{M}\) converts a point cloud \(\mathcal{P}\) into a structural representation comprising the centers and radii of medial spheres. Specifically, it yields a set of medial centers \(\mathcal{C} = \{c_1, c_2, \cdots, c_n\}\) and corresponding radii \(\mathcal{R} = \{r_1, r_2, \cdots, r_n\}\):
\begin{equation}
\mathcal{M}:\mathcal{P} \longrightarrow \langle \mathcal{C}, \mathcal{R} \rangle.
\label{eq:map}
\end{equation}

The MAT is a compact and intrinsic descriptor that captures the underlying geometry of a shape and has been widely applied in shape analysis and synthesis~\cite{amenta2001power,sun2015medial}.

\firstpara{MAT-based Adversarial Attacks.}
Given a point cloud \(\mathcal{P}\), instead of directly perturbing individual points, we propose to perturb its medial axis transform (MAT) representation \(\hat{\Theta} = \langle \mathcal{C}, \mathcal{R} \rangle\):
\begin{equation}
\setlength\abovedisplayskip{3pt}
\setlength\belowdisplayskip{3pt}
\min_{\Delta} L_{\text{mis}}(f_s, \mathcal{M}^{-1}(\hat{\Theta} + \Delta), y) + \lambda_1 D(\mathcal{P}, \mathcal{M}^{-1}(\hat{\Theta} + \Delta)),
\end{equation}
where \(\mathcal{M}^{-1}\) denotes the inverse transform that reconstructs a point cloud from its perturbed MAT representation.

By perturbing the MAT representation, the generated adversarial point clouds are expected to exhibit strong inherent adversarialness, thereby achieving improved transferability and undefendability.


\section{Method}


In this section, we introduce MAT-Adv, an adversarial attack framework that perturbs the medial axis transform (MAT) representation of point clouds to enhance transferability and undefendability. The framework integrates a deep autoencoder for MAT extraction and reconstruction, an attack module that perturbs the MAT representation, and a dropout strategy that further improves effectiveness.  Please refer to  Fig.~\ref{fig:method} for an illustration.

\begin{figure*}[!t]
    \centering
    \includegraphics[width=0.99\linewidth]{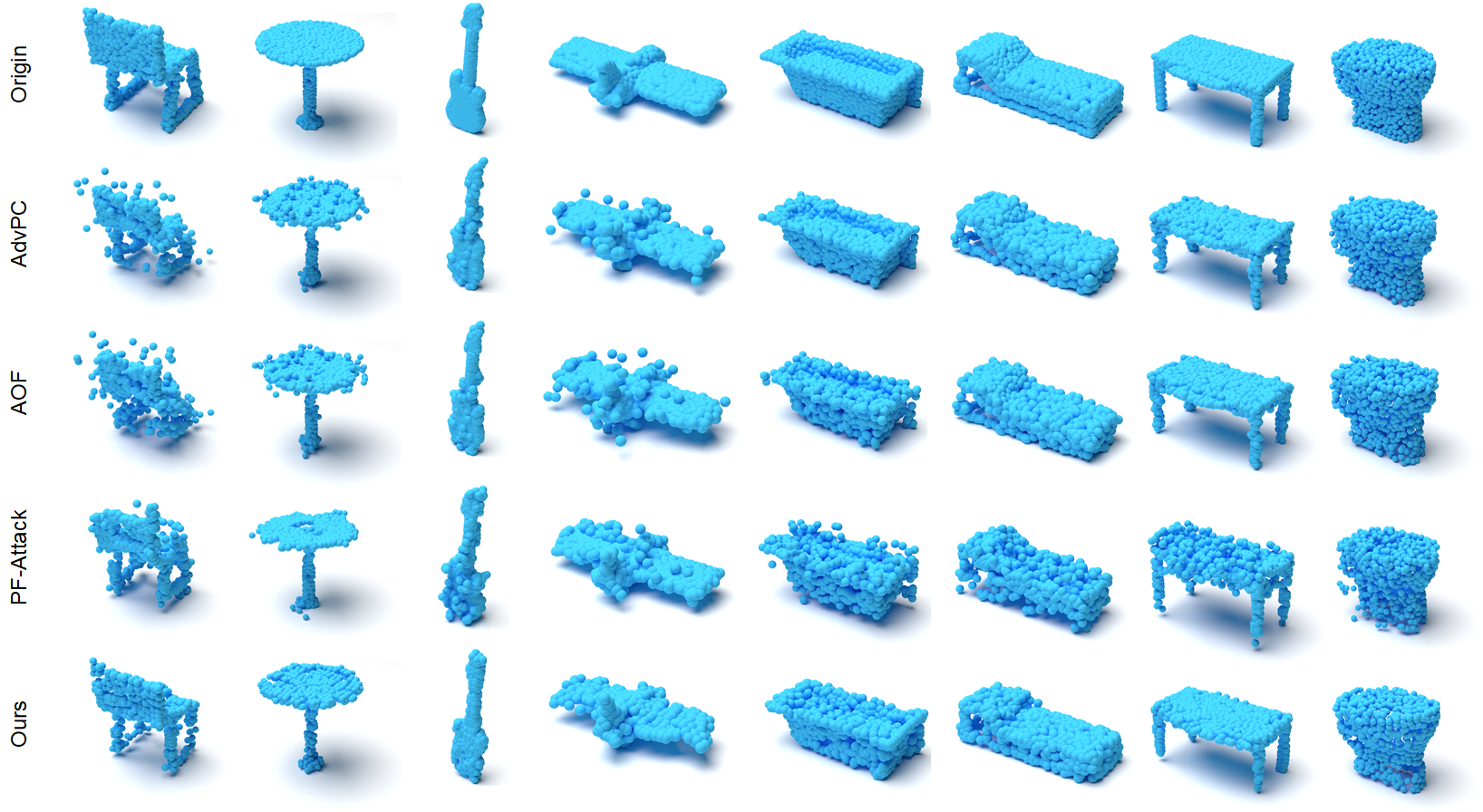}
    \vspace{-3mm}
    \caption{
    Visualizations of original and adversarial point clouds generated to fool PointNet by different attack methods
    under an $l_{\infty}$-norm perturbation budget of $\epsilon = 0.18$.
    The left four columns show examples from ShapeNet Part, and the right four columns are from ModelNet10.
 The predicted categories before and after attack from left to right are:
 {\sc{chair} $\rightarrow$ \sc{laptop}};
 {\sc{table} $\rightarrow$ \sc{lamp}};
 {\sc{guitar} $\rightarrow$ \sc{knife}};
  {\sc{airplane} $\rightarrow$ \sc{table}};
 {\sc{bathtub} $\rightarrow$ \sc{table}};
 {\sc{bed} $\rightarrow$ \sc{sofa}};
 {\sc{desk} $\rightarrow$ \sc{table}};
  {\sc{toilet} $\rightarrow$ \sc{dresser}}.
}
    \label{fig:visulization}
\end{figure*}

\subsection{Deep Medial Axis Transform}

\firstpara{Overview.}
We employ an auto-encoder to implement both the transformation \(\mathcal{M}\) and its inverse \(\mathcal{M}^{-1}\). Given a point cloud \(\mathcal{P}\) as input, the encoder \(E_{\mathcal{M}}\) predicts the MAT representation \(\Theta = \langle \mathcal{C}, \mathcal{R}, \mathcal{Z} \rangle\):
\begin{equation}
\Theta = E_{\mathcal{M}}(\mathcal{P}),
\end{equation}
where \(\mathcal{C}\) and \(\mathcal{R}\) denote the centers and radii of the medial spheres, respectively. The vector \(\mathcal{Z}\) represents auxiliary features associated with the medial spheres, facilitating a more accurate inverse transformation.

The decoder \(D_{\mathcal{M}^{-1}}\) reconstructs the original point cloud \(\mathcal{P}\) from its MAT representation \(\Theta\):
\begin{equation}
\hat{\mathcal{P}} = D_{\mathcal{M}^{-1}}(\Theta).
\end{equation}

This framework provides a compact representation of the shape’s intrinsic properties, allowing effective manipulation of the MAT representation for adversarial purposes.

\firstpara{MAT Encoder.}
Given a point cloud \(\mathcal{P} \in \mathbb{R}^{N \times 3}\), we first use PointNet++~\cite{Qi-2017-Pointnet++} to obtain sampled points \(\mathcal{P}^{s} \in \mathbb{R}^{N' \times 3}\) and their contextual features \(\mathcal{F}^{s} \in \mathbb{R}^{N' \times D_F}\), where \(N' < N\) and \(D_F\) is the feature dimension.

The encoder $E_{\mathcal M}$ then obtains the centers of the medial spheres \(\mathcal{C} \in \mathbb{R}^{n \times 3}\) and their associated features \(\mathcal{Z} \in \mathbb{R}^{n \times D_F}\) via a convex combination of the sampled points:
\begin{equation}
\begin{aligned}
\mathcal{C} &= \mathcal{W}^{\operatorname{T}} \mathcal{P}^{s}, \\
\mathcal{Z} &= \mathcal{W}^{\operatorname{T}} \mathcal{F}^{s}, 
\end{aligned}
\end{equation}
where the normalized weights \(\mathcal{W} \in \mathbb{R}^{N' \times n}\) are predicted using pointwise MLP operations on \(\mathcal{F}^{s}\) as in~\cite{lin2021point2skeleton}.

Next, we estimate the radii \(\mathcal{R} \in \mathbb{R}^{n \times 1}\) as a linear combination of the minimum distances:
\begin{equation}
\mathcal{R} = \mathcal{W}^{\operatorname{T}} \mathcal{D},
\end{equation}
where \(\mathcal{D} \in \mathbb{R}^{N' \times 1}\) contains the closest distance from each sampled point \(p^{s}_i\) to all medial sphere centers:
\[
d(p^s_i, \mathcal{C}) = \min_{c \in \mathcal{C}} \|c - p^s_i\|_2, \quad i = 1, \ldots, N'.
\]

Note that \(\mathcal{C}\), \(\mathcal{R}\), and \(\mathcal{Z}\) capture intrinsic properties of the point cloud \(\mathcal{P}\) and can be used to reconstruct \(\mathcal{P}\).

\firstpara{MAT Decoder.}
Given the predicted MAT representation \(\Theta = \langle \mathcal{C}, \mathcal{R}, \mathcal{Z} \rangle\), the MAT decoder \(D_{\mathcal{M}^{-1}}\) reconstructs the original point cloud \(\mathcal{P}\). Specifically, it uniformly samples points on each medial sphere’s surface to form an initial reconstructed point cloud \(\mathcal{Q}\) with the same number of points as \(\mathcal{P}\). For each sampled point \(q_i\), we compute its feature by identifying the \(k\) nearest medial sphere centers, whose indices form the set \(\mathcal{I}_i\). The feature \(y_{i}\) of each sampled point \(q_i\) is then computed as a weighted sum of the features  of its neighboring spheres:
\begin{equation}
y_{i} = \frac{\sum_{j \in \mathcal{I}_i} \exp{(-\|q_i - c_j\|_2)} \cdot z_j}{\sum_{j \in \mathcal{I}_i} \exp{(-\|q_i - c_j\|_2)}},
\end{equation}
where \(c_j\) and \(z_j\) represent the center and feature of the \(j\)-th medial sphere, respectively.

We further refine the reconstructed point positions, guided by the point features \(Y = [y_1, \dots, y_N]\), using
\begin{equation}
\hat{\mathcal{P}} = \operatorname{MLP}([\operatorname{MLP}([\mathcal{Q}; Y]); Y]),
\end{equation}
where \([\cdot; \cdot]\) denotes the concatenation operation.

\firstpara{MAT Training.}
We first pretrain the encoder \(E_{\mathcal M}\) following the procedure outlined in~\cite{lin2021point2skeleton}. The decoder \(D_{{\mathcal M}^{-1}}\) is then trained using a Chamfer distance loss to enforce point-wise reconstruction fidelity, along with a repulsion loss~\cite{yu2018pu} to prevent clustering of reconstructed points.

\subsection{Adversarial Attacks via Perturbating MAT}

Given an input point cloud \(\mathcal{P}\), we extract its MAT representation \(\Theta\). To generate an adversarial point cloud, we perturb the MAT representation by solving the optimization problem:
\begin{equation}
\begin{aligned}
    \min_{\Delta} & \, L_{\text{mis}}(f_s, D_{{\mathcal M}^{-1}}(\Theta + \Delta), y) \\
    & + \lambda_1 D(\mathcal{P}, D_{{\mathcal M}^{-1}}(\Theta + \Delta)) + \lambda_2 D_L(\Delta),
\end{aligned}
\label{eq:final}
\end{equation}
where \(D_L(\Delta)\) is a constraint on the MAT representation \(\Theta\), and \(\lambda_2\) is a weighting parameter.

The adversarial point cloud is then obtained by decoding this perturbed MAT representation and features:
\begin{equation}
\mathcal{P}^{\text{adv}} = D_{{\mathcal M}^{-1}}(\Theta + \Delta).
\label{eq:adv}
\end{equation}

Since the MAT representation encodes the intrinsic geometric properties of the point cloud, perturbing it enhances both the transferability and undefendability of the resulting adversarial examples across diverse models and defenses.

\subsection{Dropout Strategy on MAT Perturbation}

To mitigate overfitting to specific classifiers and avoid degradation of perturbations in the MAT representation, we introduce a dropout-style regularization inspired by Dropout~\cite{srivastava2014dropout}, which improves generalization via random deactivation.

Specifically, at each optimization iteration, we randomly mask out the perturbations on a proportion \(\rho\) of the medial spheres, keeping their positions, radii, and associated features unchanged.  
This structured masking encourages a  uniform distribution of adversarial signals and enhances both transferability and undefendability.

\section{Experimental Results}

\subsection{Experimental Setup}

\firstpara{Implementation.}  
We implement the MAT-Adv framework in PyTorch. The MAT encoder \(E_{\mathcal{M}}\) extracts 128 medial spheres, and the decoder \(D_{\mathcal{M}^{-1}}\), based on FoldingNet~\cite{yang2018foldingnet}, samples 8 points per sphere.  
It is trained for 900 epochs with the encoder frozen, followed by joint fine-tuning for 300 epochs.
The autoencoder is trained using a combination of Chamfer distance and repulsion loss~\cite{yu2018pu}, with a weight ratio of 100:1. For the repulsion loss, the number of nearest neighbors \(k\) is set to 8.  
In Eq.~\eqref{eq:final}, the geometric distortion term \(D(\cdot)\) is the Chamfer distance, and the regularization term \(D_L(\cdot)\) uses the Frobenius norm.
We set \(\lambda_1 = 10\) and \(\lambda_2 = 0.1\) for experiments with a perturbation budget of 0.18, and \(\lambda_1 = 1\), \(\lambda_2 = 0.01\) for a budget of 0.45.  
The dropout proportion \(\rho\) is fixed at 0.5.
All experiments are conducted on a workstation with dual 2.40 GHz CPUs, 128 GB RAM, and 8 NVIDIA RTX 3090 GPUs.

\begin{table*}[!t]
\centering
\caption{
Transferability performance of different attack methods on ShapeNet Part and ModelNet10.  
Transferability is measured by the attack success rate (\%) on target models using adversarial examples generated for attacking source models,  
under $l_{\infty}$-norm perturbation budgets of $\epsilon = 0.18$ and $\epsilon = 0.45$.  
Values in \textbf{bold} indicate the highest transferability, while values in \textcolor{gray}{gray} represent white-box results.
}
\centering
\label{tab:transfer_asr}
\setlength{\tabcolsep}{2.4mm}{
\scalebox{0.85}{
\centering
\begin{tabular}{@{}ccccccc|cccc@{}}
\toprule
 &
   &
   &
  \multicolumn{4}{c|}{$\epsilon=0.18$} &
  \multicolumn{4}{c}{$\epsilon=0.45$} \\ \cmidrule(l){4-11} 
\multirow{-2}{*}{Dataset} &
  \multirow{-2}{*}{Network} &
  \multirow{-2}{*}{Method} &
  PointNet &
  PointNet++ &
  DGCNN &
  PointConv &
  PointNet &
  PointNet++ &
  DGCNN &
  PointConv \\ \midrule
 &
   &
  3D-Adv &
  {\color[HTML]{747474} 100.00} &
  1.01 &
  1.32 &
  1.10 &
  {\color[HTML]{747474} 100.00} &
  1.18 &
  1.64 &
  0.17 \\
 &
   &
  AdvPC &
  {\color[HTML]{747474} 100.00} &
  14.96 &
  11.93 &
  6.75 &
  {\color[HTML]{747474} 100.00} &
  18.72 &
  14.13 &
  7.59 \\
 &
   &
  AOF &
  {\color[HTML]{747474} 100.00} &
  25.19 &
  16.98 &
  12.79 &
  {\color[HTML]{747474} 100.00} &
  33.05 &
  24.01 &
  14.86 \\
 &
   &
  PF-Attack &
  {\color[HTML]{747474} 51.25} &
  15.10 &
  10.37 &
  7.79 &
  {\color[HTML]{747474} 62.49} &
  31.52 &
  23.24 &
  17.50 \\
 &
   &
  KNN &
  {\color[HTML]{747474} 98.12} &
  17.36 &
  15.80 &
  3.97 &
  {\color[HTML]{747474} 99.34} &
  16.98 &
  19.10 &
  3.41 \\
 &
  \multirow{-6}{*}{PointNet} &
  Ours &
  {\color[HTML]{747474} 80.97} &
  \textbf{29.85} &
  \textbf{28.98} &
  \textbf{19.97} &
  {\color[HTML]{747474} 99.83} &
  \textbf{54.11} &
  \textbf{47.98} &
  \textbf{40.83} \\ \cmidrule(l){2-11} 
 &
   &
  3D-Adv &
  1.56 &
  {\color[HTML]{747474} 100.00} &
  1.45 &
  1.28 &
  2.33 &
  {\color[HTML]{747474} 100.00} &
  1.57 &
  0.35 \\
 &
   &
  AdvPC &
  16.63 &
  {\color[HTML]{747474} 99.65} &
  16.53 &
  16.91 &
  26.72 &
  {\color[HTML]{747474} 99.48} &
  23.87 &
  20.22 \\
 &
   &
  AOF &
  12.80 &
  {\color[HTML]{747474} 99.27} &
  16.95 &
  15.38 &
  21.22 &
  {\color[HTML]{747474} 99.16} &
  23.10 &
  21.02 \\
 &
   &
  PF-Attack &
  11.69 &
  {\color[HTML]{747474} 75.12} &
  10.82 &
  11.80 &
  33.92 &
  {\color[HTML]{747474} 80.20} &
  22.86 &
  21.05 \\
 &
   &
  KNN &
  5.58 &
  {\color[HTML]{747474} 100.00} &
  5.51 &
  2.05 &
  6.23 &
  {\color[HTML]{747474} 100.00} &
  5.25 &
  1.91 \\
 &
  \multirow{-6}{*}{PointNet++} &
  Ours &
  \textbf{19.62} &
  {\color[HTML]{747474} 98.90} &
  \textbf{24.56} &
  \textbf{19.74} &
  \textbf{55.36} &
  {\color[HTML]{747474} 100.00} &
  \textbf{61.19} &
  \textbf{52.77} \\ \cmidrule(l){2-11} 
 &
   &
  3D-Adv &
  2.68 &
  0.73 &
  {\color[HTML]{747474} 100.00} &
  1.25 &
  2.57 &
  0.49 &
  {\color[HTML]{747474} 100.00} &
  0.07 \\
 &
   &
  AdvPC &
  13.95 &
  \textbf{27.42} &
  {\color[HTML]{747474} 98.36} &
  15.48 &
  23.17 &
  37.47 &
  {\color[HTML]{747474} 98.89} &
  19.59 \\
 &
   &
  AOF &
  10.09 &
  24.60 &
  {\color[HTML]{747474} 100.00} &
  12.18 &
  20.49 &
  31.70 &
  {\color[HTML]{747474} 100.00} &
  16.17 \\
 &
   &
  PF-Attack &
  12.67 &
  13.50 &
  {\color[HTML]{747474} 60.02} &
  12.18 &
  39.11 &
  40.15 &
  {\color[HTML]{747474} 71.99} &
  29.54 \\
 &
   &
  KNN &
  17.95 &
  25.75 &
  {\color[HTML]{747474} 100.00} &
  12.77 &
  38.03 &
  33.72 &
  {\color[HTML]{747474} 100.00} &
  11.76 \\
 &
  \multirow{-6}{*}{DGCNN} &
  Ours &
  \textbf{21.09} &
  25.78 &
  {\color[HTML]{747474} 99.27} &
  \textbf{24.22} &
  \textbf{44.16} &
  \textbf{54.59} &
  {\color[HTML]{747474} 100.00} &
  \textbf{45.06} \\ \cmidrule(l){2-11} 
 &
   &
  3D-Adv &
  1.11 &
  1.38 &
  1.54 &
  {\color[HTML]{747474} 100.00} &
  0.42 &
  0.14 &
  2.12 &
  {\color[HTML]{747474} 100.00} \\
 &
   &
  AdvPC &
  11.55 &
  \textbf{21.89} &
  13.71 &
  {\color[HTML]{747474} 99.86} &
  18.48 &
  27.45 &
  29.97 &
  {\color[HTML]{747474} 99.72} \\
 &
   &
  AOF &
  2.30 &
  2.71 &
  2.51 &
  {\color[HTML]{747474} 100.00} &
  2.54 &
  3.03 &
  2.37 &
  {\color[HTML]{747474} 100.00} \\
 &
   &
  PF-Attack &
  11.93 &
  12.94 &
  9.78 &
  {\color[HTML]{747474} 89.32} &
  26.97 &
  30.03 &
  20.56 &
  {\color[HTML]{747474} 90.36} \\
 &
   &
  KNN &
  0.49 &
  0.87 &
  1.67 &
  {\color[HTML]{747474} 100.00} &
  0.56 &
  0.66 &
  2.30 &
  {\color[HTML]{747474} 100.00} \\
\multirow{-24}{*}{\rotatebox{90}{ShapeNet Part}} &
  \multirow{-6}{*}{PointConv} &
  Ours &
  \textbf{12.58} &
  13.15 &
  \textbf{19.83} &
  {\color[HTML]{747474} 90.74} &
  \textbf{35.32} &
  \textbf{37.13} &
  \textbf{39.32} &
  {\color[HTML]{747474} 97.98} \\ \midrule
 &
   &
  3D-Adv &
  {\color[HTML]{808080} 99.56} &
  1.54 &
  1.36 &
  1.43 &
  {\color[HTML]{808080} 100.00} &
  1.21 &
  1.43 &
  1.32 \\
 &
   &
  AdvPC &
  {\color[HTML]{808080} 98.24} &
  7.49 &
  5.18 &
  8.99 &
  {\color[HTML]{808080} 99.89} &
  8.70 &
  5.18 &
  8.88 \\
 &
   &
  AOF &
  {\color[HTML]{808080} 97.88} &
  23.02 &
  8.81 &
  24.27 &
  {\color[HTML]{808080} 99.42} &
  24.56 &
  9.47 &
  23.24 \\
 &
   &
  PF-Attack &
  {\color[HTML]{808080} 77.75} &
  23.87 &
  19.05 &
  \textbf{30.84} &
  {\color[HTML]{808080} 83.81} &
  26.54 &
  20.93 &
  39.76 \\
 &
   &
  KNN &
  {\color[HTML]{808080} 90.31} &
  4.30 &
  7.16 &
  4.30 &
  {\color[HTML]{808080} 90.53} &
  5.18 &
  6.72 &
  5.51 \\
 &
  \multirow{-6}{*}{PointNet} &
  Ours &
  {\color[HTML]{808080} 81.57} &
  \textbf{24.96} &
  \textbf{24.78} &
  28.09 &
  {\color[HTML]{808080} 88.77} &
  \textbf{47.47} &
  \textbf{49.56} &
  \textbf{51.94} \\ \cmidrule(l){2-11} 
 &
   &
  3D-Adv &
  3.12 &
  {\color[HTML]{808080} 100.00} &
  1.04 &
  1.56 &
  2.76 &
  {\color[HTML]{808080} 100.00} &
  1.80 &
  2.16 \\
 &
   &
  AdvPC &
  9.91 &
  {\color[HTML]{808080} 94.71} &
  8.37 &
  20.03 &
  10.59 &
  {\color[HTML]{808080} 93.83} &
  9.15 &
  24.07 \\
 &
   &
  AOF &
  18.17 &
  {\color[HTML]{808080} 98.13} &
  11.78 &
  33.59 &
  20.59 &
  {\color[HTML]{808080} 98.13} &
  13.22 &
  35.57 \\
 &
   &
  PF-Attack &
  20.15 &
  {\color[HTML]{808080} 91.52} &
  23.90 &
  \textbf{39.98} &
  38.44 &
  {\color[HTML]{808080} 94.06} &
  32.81 &
  50.25 \\
 &
   &
  KNN &
  7.71 &
  {\color[HTML]{808080} 100.00} &
  10.13 &
  23.57 &
  9.58 &
  {\color[HTML]{808080} 100.00} &
  11.45 &
  24.89 \\
 &
  \multirow{-6}{*}{PonitNet++} &
  Ours &
  \textbf{21.58} &
  {\color[HTML]{808080} 91.74} &
  \textbf{28.26} &
  30.73 &
  \textbf{47.58} &
  {\color[HTML]{808080} 98.24} &
  \textbf{61.12} &
  \textbf{67.73} \\ \cmidrule(l){2-11} 
 &
   &
  3D-Adv &
  1.43 &
  1.32 &
  {\color[HTML]{808080} 100.00} &
  0.66 &
  1.53 &
  1.76 &
  {\color[HTML]{808080} 100.00} &
  0.97 \\
 &
   &
  AdvPC &
  13.99 &
  19.16 &
  {\color[HTML]{808080} 90.97} &
  21.04 &
  15.20 &
  21.37 &
  {\color[HTML]{808080} 92.84} &
  22.69 \\
 &
   &
  AOF &
  23.90 &
  \textbf{33.81} &
  {\color[HTML]{808080} 99.89} &
  31.50 &
  36.12 &
  41.08 &
  {\color[HTML]{808080} 100.00} &
  36.26 \\
 &
   &
  PF-Attack &
  17.07 &
  23.57 &
  {\color[HTML]{808080} 82.82} &
  31.39 &
  35.79 &
  29.52 &
  {\color[HTML]{808080} 85.02} &
  41.19 \\
 &
   &
  KNN &
  9.47 &
  8.37 &
  {\color[HTML]{808080} 100.00} &
  13.33 &
  10.46 &
  9.25 &
  {\color[HTML]{808080} 100.00} &
  13.66 \\
 &
  \multirow{-6}{*}{DGCNN} &
  Ours &
  \textbf{24.13} &
  24.34 &
  {\color[HTML]{808080} 97.14} &
  \textbf{40.75} &
  \textbf{75.22} &
  \textbf{83.59} &
  {\color[HTML]{808080} 100.00} &
  \textbf{73.13} \\ \cmidrule(l){2-11} 
 &
   &
  3D-Adv &
  1.12 &
  1.45 &
  1.56 &
  {\color[HTML]{808080} 100.00} &
  1.32 &
  1.76 &
  2.33 &
  {\color[HTML]{808080} 100.00} \\
 &
   &
  AdvPC &
  5.70 &
  10.01 &
  5.70 &
  {\color[HTML]{808080} 99.78} &
  6.25 &
  10.24 &
  6.25 &
  {\color[HTML]{808080} 100.00} \\
 &
   &
  AOF &
  4.86 &
  9.14 &
  7.05 &
  {\color[HTML]{808080} 100.00} &
  5.63 &
  8.81 &
  5.84 &
  {\color[HTML]{808080} 100.00} \\
 &
   &
  PF-Attack &
  12.00 &
  \textbf{26.32} &
  18.06 &
  {\color[HTML]{808080} 96.15} &
  19.32 &
  30.68 &
  19.89 &
  {\color[HTML]{808080} 95.45} \\
 &
   &
  KNN &
  3.85 &
  3.85 &
  4.74 &
  {\color[HTML]{808080} 100.00} &
  3.85 &
  3.41 &
  4.41 &
  {\color[HTML]{808080} 100.00} \\
\multirow{-24}{*}{\rotatebox{90}{ModelNet 10}} &
  \multirow{-6}{*}{PointConv} &
  Ours &
  \textbf{13.89} &
  22.14 &
  \textbf{21.26} &
  {\color[HTML]{808080} 97.03} &
  \textbf{35.90} &
  \textbf{56.17} &
  \textbf{48.24} &
  {\color[HTML]{808080} 98.13} \\ \bottomrule
\end{tabular}}
}
\end{table*}

\firstpara{Datasets.}  
We evaluate our approach on three publicly available datasets, including the synthetic datasets ModelNet10~\cite{mn10} and ShapeNet Part~\cite{yi-2016-shapenet}, and the real-world scanned dataset ScanObjectNN~\cite{uy2019revisiting}.

ShapeNet Part consists of 14,007 training and 2,874 testing point clouds from 16 categories, while ModelNet10 includes 4,899 shapes from 10 categories with 3,991 for training and 908 for testing. ScanObjectNN is a real-world scanned dataset that contains 15,000 objects extracted from real-world scans, which are categorized into 15 classes. Due to the existence of background, noise, and occlusions, this benchmark poses significant challenges to existing point cloud analysis methods. Each point cloud is randomly sampled to 1,024 points, following the protocol in~\cite{Xiang-2019-Generating}. 

\begin{table*}[!t]
\centering
\caption{Transferability performance of different attack methods on ScanObjectNN.  
Transferability is measured by the attack success rate (\%) on target models using adversarial examples generated for attacking source models,  
under $l_{\infty}$-norm perturbation budgets of $\epsilon = 0.18$ and $\epsilon = 0.45$.  
Values in \textbf{bold} indicate the highest transferability, while values in \textcolor{gray}{gray} represent white-box results.}
\label{tab:transferscan}
\setlength{\tabcolsep}{2.9mm}{
\scalebox{0.89}{
\begin{tabular}{@{}cccccc|cccc@{}}
\toprule
 &
   &
  \multicolumn{4}{c|}{$\epsilon=0.18$} &
  \multicolumn{4}{c}{$\epsilon=0.45$} \\ \cmidrule(l){3-10} 
\multirow{-2}{*}{Network} &
  \multirow{-2}{*}{Method} &
  PointNet &
  PointNet++ &
  DGCNN &
  PointConv &
  PointNet &
  PointNet++ &
  DGCNN &
  PointConv \\ \midrule
 &
  AdvPC &
  {\color[HTML]{ADADAD} 100.00} &
  27.71 &
  26.33 &
  33.22 &
  {\color[HTML]{ADADAD} 100.00} &
  26.68 &
  25.13 &
  34.60 \\
 &
  AOF &
  {\color[HTML]{ADADAD} 100.00} &
  25.30 &
  23.58 &
  32.36 &
  {\color[HTML]{ADADAD} 100.00} &
  23.75 &
  23.24 &
  32.53 \\
 &
  PF-Attack &
  {\color[HTML]{ADADAD} 100.00} &
  45.47 &
  39.05 &
  \textbf{67.55} &
  {\color[HTML]{ADADAD} 100.00} &
  49.99 &
  41.46 &
  \textbf{66.61} \\
\multirow{-4}{*}{PointNet} &
  Ours &
  {\color[HTML]{ADADAD} 99.83} &
  \textbf{49.91} &
  \textbf{44.92} &
  51.29 &
  {\color[HTML]{ADADAD} 99.66} &
  \textbf{50.95} &
  \textbf{46.99} &
  59.74 \\ \midrule
 &
  AdvPC &
  12.91 &
  {\color[HTML]{ADADAD} 98.80} &
  25.13 &
  36.49 &
  12.05 &
  {\color[HTML]{ADADAD} 98.28} &
  28.06 &
  36.14 \\
 &
  AOF &
  19.49 &
  {\color[HTML]{ADADAD} 96.90} &
  34.25 &
  45.44 &
  18.93 &
  {\color[HTML]{ADADAD} 97.59} &
  36.32 &
  46.64 \\
 &
  PF-Attack &
  \textbf{37.52} &
  {\color[HTML]{ADADAD} 100.00} &
  46.60 &
  50.06 &
  \textbf{41.14} &
  {\color[HTML]{ADADAD} 100.00} &
  45.88 &
  55.92 \\
\multirow{-4}{*}{PointNet++} &
  Ours &
  27.73 &
  {\color[HTML]{ADADAD} 100.00} &
  \textbf{49.72} &
  \textbf{55.94} &
  30.28 &
  {\color[HTML]{ADADAD} 97.96} &
  \textbf{47.68} &
  \textbf{59.04} \\ \midrule
 &
  AdvPC &
  17.21 &
  35.80 &
  {\color[HTML]{ADADAD} 96.90} &
  45.78 &
  15.83 &
  38.04 &
  {\color[HTML]{ADADAD} 98.11} &
  48.54 \\
 &
  AOF &
  22.55 &
  41.65 &
  {\color[HTML]{ADADAD} 100.00} &
  46.99 &
  22.41 &
  41.31 &
  {\color[HTML]{ADADAD} 100.00} &
  49.05 \\
 &
  PF-Attack &
  \textbf{41.20} &
  60.27 &
  {\color[HTML]{ADADAD} 100.00} &
  60.72 &
  \textbf{45.87} &
  61.23 &
  {\color[HTML]{ADADAD} 100.00} &
  67.42 \\
\multirow{-4}{*}{DGCNN} &
  Ours &
  22.72 &
  \textbf{73.35} &
  {\color[HTML]{ADADAD} 100.00} &
  \textbf{67.99} &
  23.41 &
  \textbf{71.26} &
  {\color[HTML]{ADADAD} 100.00} &
  \textbf{68.69} \\ \midrule
 &
  AdvPC &
  5.51 &
  16.52 &
  14.63 &
  {\color[HTML]{ADADAD} 98.80} &
  4.99 &
  15.83 &
  17.04 &
  {\color[HTML]{ADADAD} 99.14} \\
 &
  AOF &
  2.93 &
  15.66 &
  12.91 &
  {\color[HTML]{ADADAD} 100.00} &
  5.16 &
  13.08 &
  13.94 &
  {\color[HTML]{ADADAD} 100.00} \\
 &
  PF-Attack &
  \textbf{35.44} &
  32.98 &
  26.32 &
  {\color[HTML]{ADADAD} 100.00} &
  \textbf{37.50} &
  36.87 &
  30.63 &
  {\color[HTML]{ADADAD} 100.00} \\
\multirow{-4}{*}{PointConv} &
  Ours &
  10.50 &
  \textbf{39.76} &
  \textbf{30.98} &
  {\color[HTML]{ADADAD} 100.00} &
  11.53 &
  \textbf{40.62} &
  \textbf{32.70} &
  {\color[HTML]{ADADAD} 100.00} \\ \bottomrule
\end{tabular}}}
\end{table*}

\firstpara{Victim DNN Classifiers.}
We utilize four commonly used deep learning models as victim classifiers: the MLP-based PointNet~\cite{Qi-2017-Pointnet}, the hierarchical PointNet++~\cite{Qi-2017-Pointnet++}, the graph-based DGCNN~\cite{Wang-2019-DGCNN}, and the convolutional PointConv~\cite{Wu-2019-Pointconv}. Moreover, as advanced models typically increase the difficulty of attacks, we also evaluate the transferability of our proposed attack on the state-of-the-art target models, i.e., CurveNet\cite{xiang2021walk}, PCT\cite{guo2021pct}, PT\cite{zhao2021PT}, and Mamba3D\cite{han2024mamba3d}. Each model is trained following the procedures specified in their original publications.

\firstpara{Baselines.}
We select five state-of-the-art point cloud adversarial attack methods as baselines for comparison: 3D-Adv~\cite{Xiang-2019-Generating}, AdvPC~\cite{hamdi2020advpc}, AOF~\cite{liu2022boosting}, PF-Attack~\cite{he2023generating}, and KNN~\cite{tsai-2020-robust(smooth)}.

\begin{table*}[!b]
\centering
\caption{Attack transferability on state-of-the-art models on ShapeNet Part dataset. Measure performance in terms of attack success rate (\%). Adversarial point clouds are generated on DGCNN. Values in \textbf{bold} indicate the highest transferability, while values in \textcolor{gray}{gray} represent white-box results.}
\label{tab:sota_transfer}
\setlength{\tabcolsep}{3.6mm}{
\scalebox{0.89}{
\begin{tabular}{@{}cccccc|ccccc@{}}
\toprule
                         & \multicolumn{5}{c|}{$\epsilon=0.18$}                                  & \multicolumn{5}{c}{$\epsilon=0.45$}                                   \\ \cmidrule(l){2-11} 
\multirow{-2}{*}{Method} & DGCNN                         & CurveNet & PCT   & PT    & Mamba3D & DGCNN                         & CurveNet & PCT   & PT    & Mamba3D \\ \midrule
AdvPC                    & {\color[HTML]{ADADAD} 99.10}  & 12.66    & 12.52 & 15.90 & 52.17   & {\color[HTML]{ADADAD} 98.96}  & 23.83    & 17.67 & 18.82 & 59.13   \\
AOF                      & {\color[HTML]{ADADAD} 100.00} & 11.03    & 8.87  & 12.00 & 43.93   & {\color[HTML]{ADADAD} 100.00} & 19.41    & 16.56 & 18.57 & 51.17   \\
PF-Attack &
  {\color[HTML]{ADADAD} 83.44} &
  9.32 &
  10.99 &
  12.10 &
  33.67 &
  {\color[HTML]{ADADAD} 94.43} &
  \textbf{29.50} &
  27.55 &
  \textbf{34.16} &
  48.66 \\
Ours &
  {\color[HTML]{ADADAD} 95.10} &
   \textbf{15.76} &
  \textbf{21.77} &
  \textbf{16.42} &
  \textbf{60.07} &
  {\color[HTML]{ADADAD} 99.09} &
  24.73 &
  \textbf{29.53} &
  23.64 &
  \textbf{72.92} \\ \bottomrule
\end{tabular}}}
\end{table*}

\firstpara{Evaluation Setting and Metrics.}
To evaluate the effectiveness of different methods, we use the attack success rate (ASR), defined as the percentage of the misclassified adversarial samples out of all the clean samples that are classified correctly to evaluate the transferability. Experiments are conducted under two $l_{\infty}$-norm perturbation budgets: $\epsilon = 0.18$ and $\epsilon = 0.45$.



\begin{figure*}[!t]
    \centering
    \includegraphics[width=0.91\linewidth]{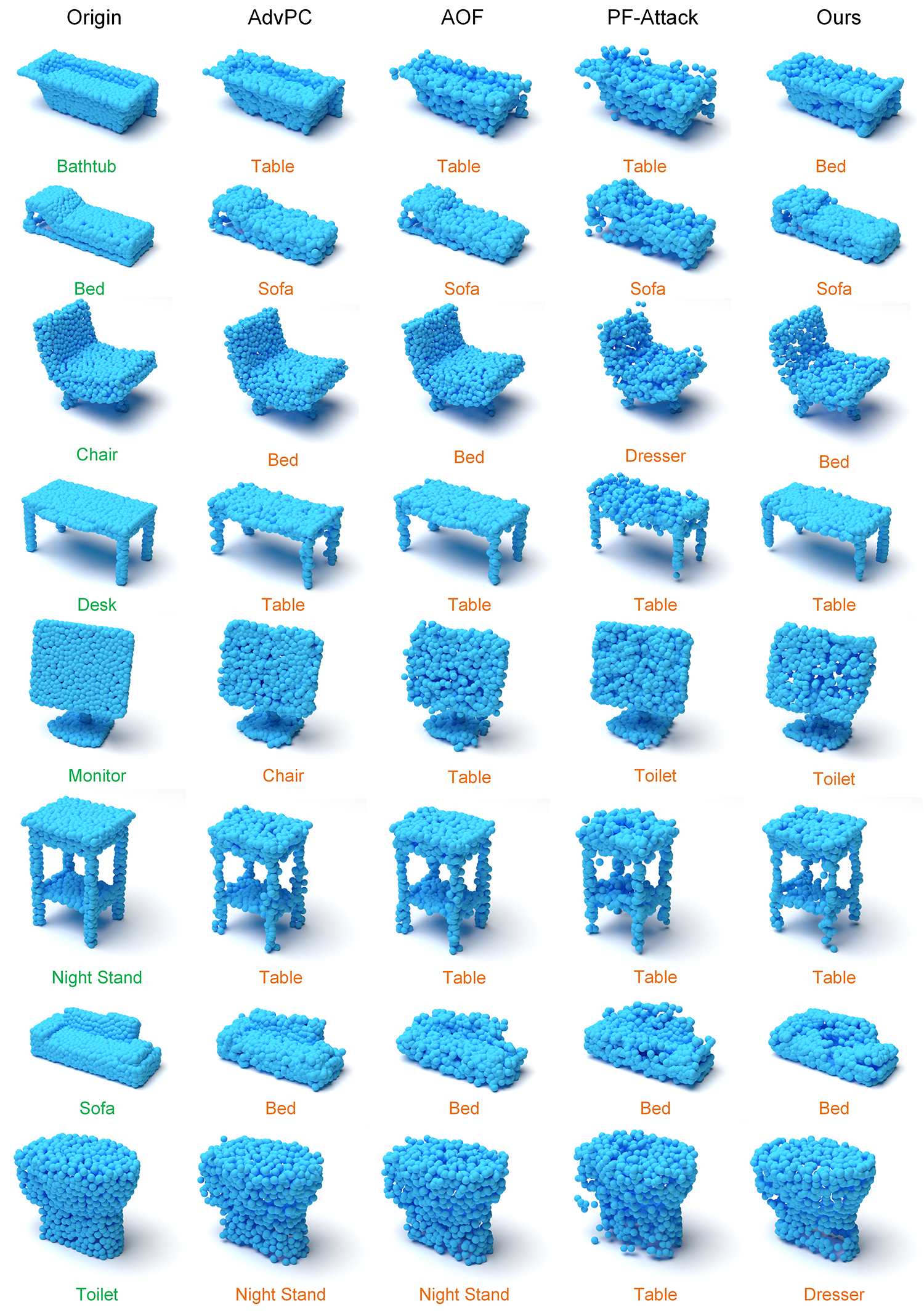}
    \caption{Visualizations of original and adversarial point clouds generated to fool PointNet by different attack methods
    under an $l_{\infty}$-norm perturbation budget of $\epsilon = 0.18$.
    The top four rows show examples from ModelNet10, and the bottom four rows present examples from ShapeNet Part.
 The predicted categories before and after attack from top to bottom are:
 {\sc{chair} $\rightarrow$ \sc{bed}};
 {\sc{monitor} $\rightarrow$ \sc{toilet}};
 {\sc{night stand} $\rightarrow$ \sc{table}};
  {\sc{sofa} $\rightarrow$ \sc{bed}};
 {\sc{car} $\rightarrow$ \sc{rocket}};
 {\sc{motorbike} $\rightarrow$ \sc{rocket}};
 {\sc{skateboard} $\rightarrow$ \sc{rocket}};
  {\sc{lamp} $\rightarrow$ \sc{earphone}}.
}
    \label{fig:visulization_2}
\end{figure*}

\firstpara{Qualitative Evaluation.}  
We present visualizations of adversarial point clouds generated by different attack methods on the ShapeNet Part and ModelNet10 in Fig.~\ref{fig:visulization}.  
The results show that our method, which perturbs the medial spheres, leads to localized and subtle surface deformations while preserving the overall shape structure.  
By manipulating the intrinsic geometric representation, our approach ensures that perturbations are spatially coherent and structurally plausible.  
In contrast, point-based and frequency-domain perturbation methods often introduce noticeable outliers or scattered distortions.  
The visual differences highlight the advantages of our method in producing more natural adversarial shapes, which contribute to stronger transferability and increased robustness against a variety of defense strategies.
Additional visualization results are provided in Fig.\ref{fig:visulization_2}.

\begin{table*}[!t]
\centering
\caption{Attack success rate (\%) of different methods with and without defenses under an $l_{\infty}$-norm perturbation budget of $\epsilon = 0.45$ on ShapeNet Part and ModelNet10. The best results are highlighted in \textbf{bold}, and the second-best are \underline{underlined}.}
\label{tab:defense_main}
\setlength{\tabcolsep}{2.2mm}{
\scalebox{0.86}{
\begin{tabular}{@{}cccccccc|cccccc@{}}
\toprule
\multirow{2}{*}{Network} &
  \multirow{2}{*}{Method} &
  \multicolumn{6}{c|}{ShapeNet Part} &
  \multicolumn{6}{c}{ModelNet 10} \\ \cmidrule(l){3-14} 
 &
   &
  - &
  SRS &
  SOR &
  DUP-Net &
  IF-Defense &
  AT &
  - &
  SRS &
  SOR &
  DUP-Net &
  IF-Defense &
  AT \\ \midrule
\multirow{6}{*}{{\rotatebox{90}{PointNet}}} &
  3D-Adv &
  100.00 &
  6.30 &
  0.38 &
  0.59 &
  0.59 &
  6.75 &
  100.00 &
  21.04 &
  5.73 &
  4.30 &
  2.53 &
  3.08 \\
 &
  AdvPC &
  100.00 &
  81.43 &
  10.78 &
  7.44 &
  1.18 &
  {\ul 58.78} &
  99.89 &
  80.53 &
  40.09 &
  25.44 &
  8.26 &
  45.93 \\
 &
  AOF &
  100.00 &
  \textbf{96.21} &
  39.39 &
  {\ul 32.53} &
  7.10 &
  \textbf{64.56} &
  99.42 &
  \textbf{88.66} &
  \textbf{81.72} &
  {\ul 74.45} &
  26.98 &
  {\ul 48.79} \\
 &
  PF-Attack &
  62.49 &
  49.27 &
  27.07 &
  27.70 &
  {\ul 17.64} &
  11.20 &
  83.81 &
  67.07 &
  59.69 &
  55.62 &
  {\ul 41.85} &
  5.73 \\
 &
  KNN &
  99.34 &
  23.30 &
  {\ul 70.04} &
  16.00 &
  5.08 &
  13.95 &
  90.53 &
  65.09 &
  38.00 &
  23.79 &
  7.93 &
  5.62 \\
 &
  Ours &
  100.00 &
  {\ul 90.43} &
  \textbf{71.55} &
  \textbf{63.65} &
  \textbf{41.95} &
  29.08 &
  88.77 &
  {\ul 82.93} &
  {\ul 76.21} &
  \textbf{75.44} &
  \textbf{55.62} &
  \textbf{52.93} \\ \midrule
\multirow{6}{*}{{\rotatebox{90}{PointNet++}}} &
  3D-Adv &
  100.00 &
  4.42 &
  1.45 &
  1.35 &
  1.38 &
  7.65 &
  100.00 &
  27.15 &
  8.01 &
  5.27 &
  2.34 &
  8.15 \\
 &
  AdvPC &
  99.45 &
  {\ul 82.40} &
  54.64 &
  45.25 &
  10.85 &
  {\ul 26.96} &
  93.83 &
  63.00 &
  56.39 &
  35.46 &
  10.46 &
  {\ul 18.94} \\
 &
  AOF &
  99.16 &
  78.92 &
  {\ul 63.23} &
  {\ul 56.90} &
  {\ul 22.37} &
  26.89 &
  98.13 &
  77.53 &
  79.19 &
  {\ul 65.31} &
  29.85 &
  18.06 \\
 &
  PF-Attack &
  80.20 &
  46.56 &
  32.81 &
  30.78 &
  15.78 &
  19.43 &
  94.06 &
  68.17 &
  62.56 &
  62.33 &
  {\ul 48.68} &
  15.93 \\
 &
  KNN &
  99.11 &
  59.15 &
  38.67 &
  12.05 &
  1.28 &
  6.88 &
  100.00 &
  {\ul 87.33} &
  \textbf{91.96} &
  63.99 &
  17.73 &
  10.13 \\
 &
  Ours &
  100.00 &
  \textbf{93.46} &
  \textbf{77.08} &
  \textbf{63.83} &
  \textbf{37.01} &
  \textbf{29.02} &
  98.24 &
  \textbf{90.86} &
  {\ul 90.97} &
  \textbf{87.33} &
  \textbf{75.44} &
  \textbf{25.95} \\ \midrule
\multirow{6}{*}{{\rotatebox{90}{DGCNN}}} &
  3D-Adv &
  100.00 &
  2.92 &
  1.77 &
  4.59 &
  0.70 &
  1.91 &
  100.00 &
  8.17 &
  6.49 &
  20.19 &
  2.76 &
  6.39 \\
 &
  AdvPC &
  98.89 &
  62.33 &
  {\ul 49.04} &
  40.38 &
  10.47 &
  34.82 &
  92.84 &
  31.17 &
  38.99 &
  39.87 &
  8.70 &
  21.04 \\
 &
  AOF &
  100.00 &
  53.25 &
  48.28 &
  41.32 &
  13.04 &
  \textbf{41.74} &
  100.00 &
  {\ul 56.83} &
  57.49 &
  49.67 &
  21.59 &
  {\ul 25.55} \\
 &
  PF-Attack &
  71.99 &
  39.93 &
  30.64 &
  40.70 &
  {\ul 19.48} &
  8.90 &
  85.02 &
  47.36 &
  55.29 &
  {\ul 57.49} &
  {\ul 36.89} &
  19.93 \\
 &
  KNN &
  100.00 &
  {\ul 73.18} &
  82.16 &
  {\ul 46.19} &
  18.19 &
  5.95 &
  100.00 &
  44.93 &
  {\ul 68.28} &
  40.86 &
  14.87 &
  6.06 \\
 &
  Ours &
  100.00 &
  \textbf{86.75} &
  \textbf{83.55} &
  \textbf{66.33} &
  \textbf{41.91} &
  {\ul 35.95} &
  100.00 &
  \textbf{85.68} &
  \textbf{87.33} &
  \textbf{62.00} &
  \textbf{56.94} &
  \textbf{28.75} \\ \midrule
\multirow{6}{*}{{\rotatebox{90}{PointConv}}} &
  3D-Adv &
  100.00 &
  2.92 &
  1.77 &
  4.59 &
  0.70 &
  9.57 &
  100.00 &
  20.16 &
  22.03 &
  18.59 &
  6.09 &
  10.68 \\
 &
  AdvPC &
  98.89 &
  62.33 &
  {\ul 49.04} &
  40.38 &
  10.47 &
  {\ul 57.67} &
  100.00 &
  {\ul 80.18} &
  {\ul 91.52} &
  62.33 &
  44.27 &
  44.27 \\
 &
  AOF &
  100.00 &
  53.25 &
  48.28 &
  41.32 &
  13.04 &
  \textbf{60.35} &
  100.00 &
  77.31 &
  90.09 &
  61.01 &
  46.04 &
  {\ul 44.82} \\
 &
  PF-Attack &
  71.99 &
  39.93 &
  30.64 &
  40.70 &
  {\ul 19.48} &
  20.17 &
  95.45 &
  73.13 &
  81.72 &
  {\ul 64.63} &
  {\ul 58.81} &
  13.77 \\
 &
  KNN &
  100.00 &
  {\ul 73.18} &
  82.16 &
  {\ul 46.19} &
  18.19 &
  14.78 &
  100.00 &
  47.08 &
  \textbf{91.63} &
  40.20 &
  13.55 &
  12.22 \\
 &
  Ours &
  100.00 &
  \textbf{86.75} &
  \textbf{83.55} &
  \textbf{66.33} &
  \textbf{41.91} &
  44.76 &
  98.13 &
  \textbf{87.00} &
  89.76 &
  \textbf{82.71} &
  \textbf{82.82} &
  \textbf{51.26} \\ \bottomrule
\end{tabular}}}
\end{table*}

\begin{table*}[!h]
\centering
\caption{Transferability performance under defenses. Adversarial examples are generated using PointNet as the surrogate model on the ShapeNet Part dataset under an $l_{\infty}$-norm perturbation budget of $\epsilon = 0.45$, and then evaluated on other models under various defenses.}
\label{tab: transfer_defense}
\setlength{\tabcolsep}{3.2mm}{
\scalebox{0.86}{
\begin{tabular}{@{}c|cccc|c|cccc@{}}
\toprule
Defense              & Attack    & PointNet++ & DGCNN & PointConv & Defense                     & Atttack   & PointNet++ & DGCNN & PointConv \\ \midrule
\multirow{4}{*}{SRS} & AdvPC     & 11.55      & 5.39  & 4.07      & \multirow{4}{*}{DUP-Net}    & AdvPC     & 6.57       & 11.90 & 4.45      \\
                     & AOF       & 23.79      & 13.15 & 8.38      &                             & AOF       & 5.57       & 23.51 & 8.80      \\
                     & PF-Attack & 20.83      & 11.1  & 8.24      &                             & PF-Attack & 7.27       & 23.25 & 7.79      \\
 & Ours & \textbf{34.19} & \textbf{26.19} & \textbf{11.44} &  & Ours & \textbf{17.91} & \textbf{30.89} & \textbf{16.17} \\ \midrule
\multirow{4}{*}{SOS} & AdvPC     & 1.36       & 1.84  & 4.77      & \multirow{4}{*}{IF-Defense} & AdvPC     & 0.52       & 1.04  & 0.83      \\
                     & AOF       & 5.25       & 6.23  & 2.89      &                             & AOF       & 2.82       & 2.92  & 1.84      \\
                     & PF-Attack & 6.26       & 6.72  & 2.26      &                             & PF-Attack & 4.52       & 4.24  & 1.50      \\
 & Ours & \textbf{17.7}  & \textbf{17.11} & \textbf{9.43}  &  & Ours & \textbf{12.52} & \textbf{10.57} & \textbf{11.15} \\ \bottomrule
\end{tabular}
}}
\end{table*}

\firstpara{Performance on Attack and Transferability.}  
As summarized in Tab.~\ref{tab:transfer_asr}, most methods achieve high attack success rates (ASR) under white-box settings, with several reaching 100\%.  
However, their performance degrades significantly in transfer scenarios. For example, 3D-Adv exhibits a dramatic drop, with ASR falling to as low as 1\% in most cases.  
Even methods with relatively better transferability, such as AdvPC, may experience a decline below 10\% under $\epsilon = 0.18$, especially when transferring from PointNet to PointConv.  
While increasing the perturbation budget to $\epsilon = 0.45$ leads to some improvement, the gains remain limited for most methods.
Among the baselines, AOF—a frequency-domain method—and PF-Attack—which uses random factorization—perform comparatively better, maintaining ASR around 20\% in some transfer settings.  
In contrast, our MAT-Adv, which perturbs intrinsic structural representations, consistently achieves the highest transfer ASR across the majority of configurations, confirming  its superiority.

To further evaluate transferability under realistic scenarios, we conduct experiments on the ScanObjectNN dataset, as shown in Tab.~\ref{tab:transferscan}.
Compared to the synthetic datasets, all methods exhibit noticeably higher transfer attack success rates on this real-world dataset.
However, this apparent improvement does not necessarily indicate stronger attack transferability. Instead, it primarily results from the degraded performance of point cloud classifiers when confronted with more complex and noisy, and incomplete input in ScanObjectNN.
Such degraded robustness reduces the overall decision margin of the models, making them inherently more vulnerable to even weak adversarial perturbations.
For instance, under $\epsilon = 0.18$, methods like AdvPC and AOF achieve moderate ASRs.
Nonetheless, our method, MAT-Adv, still consistently surpasses all baselines across nearly all configurations. 

We also investigate the transferability of adversarial examples to state-of-the-art architectures beyond the original source models.
In this setting, adversarial point clouds are generated on DGCNN and tested on four diverse architectures: CurveNet, PCT, Point Transformer (PT), and Mamba3D, as detailed in Tab.~\ref{tab:sota_transfer}.
These results confirm that our method not only transfers well within conventional architectures, but also remains robust when facing structurally and functionally distinct models.

\firstpara{Performance on Undefendability.}
We evaluate the resistance of different attack methods against five defense strategies: simple random sampling (SRS), which randomly drops 500 points; statistical outlier removal (SOR), which removes points based on statistical deviation from local neighbors; DUP-Net~\cite{zhou-2019-dup}, an upsampling method incorporating SOR; IF-Defense~\cite{wu2020if}, which combines outlier filtering with shape optimization; and Adversarial Training~\cite{liu-2019-extending}, which enhances the generalization ability of the classifier by incorporating adversarial point clouds into the training process of the target model.

As shown in Tab.~\ref{tab:defense_main}, our method—which perturbs intrinsic structural properties—demonstrates strong resilience against all four defenses, consistently outperforming baseline attacks by a large margin in most settings.  
For example, on PointNet trained with ShapeNet Part, IF-Defense reduces the ASR of most methods to around 10\%, whereas our MAT-Adv maintains over 40\% ASR under the same setting.  
These results highlight the robustness of our approach against a variety of defense mechanisms.

Besides, we also conduct additional experiments to assess the transferability of different attack methods under four defense strategies: SRS, SOR, DUP-Net~\cite{zhou-2019-dup}, and IF-Defense~\cite{wu2020if}.

As shown in Tab.~\ref{tab: transfer_defense}, our method consistently achieves the highest transfer attack success rates across most defense scenarios, demonstrating strong resilience and the ability to maintain transferability even under adversarial defenses.
This highlights the advantage of perturbing intrinsic representations, which are less vulnerable to common defense mechanisms.

\subsection{Ablation Studies and Other Analysis}

\begin{figure}[]
    \centering

    \includegraphics[width=0.72\linewidth]{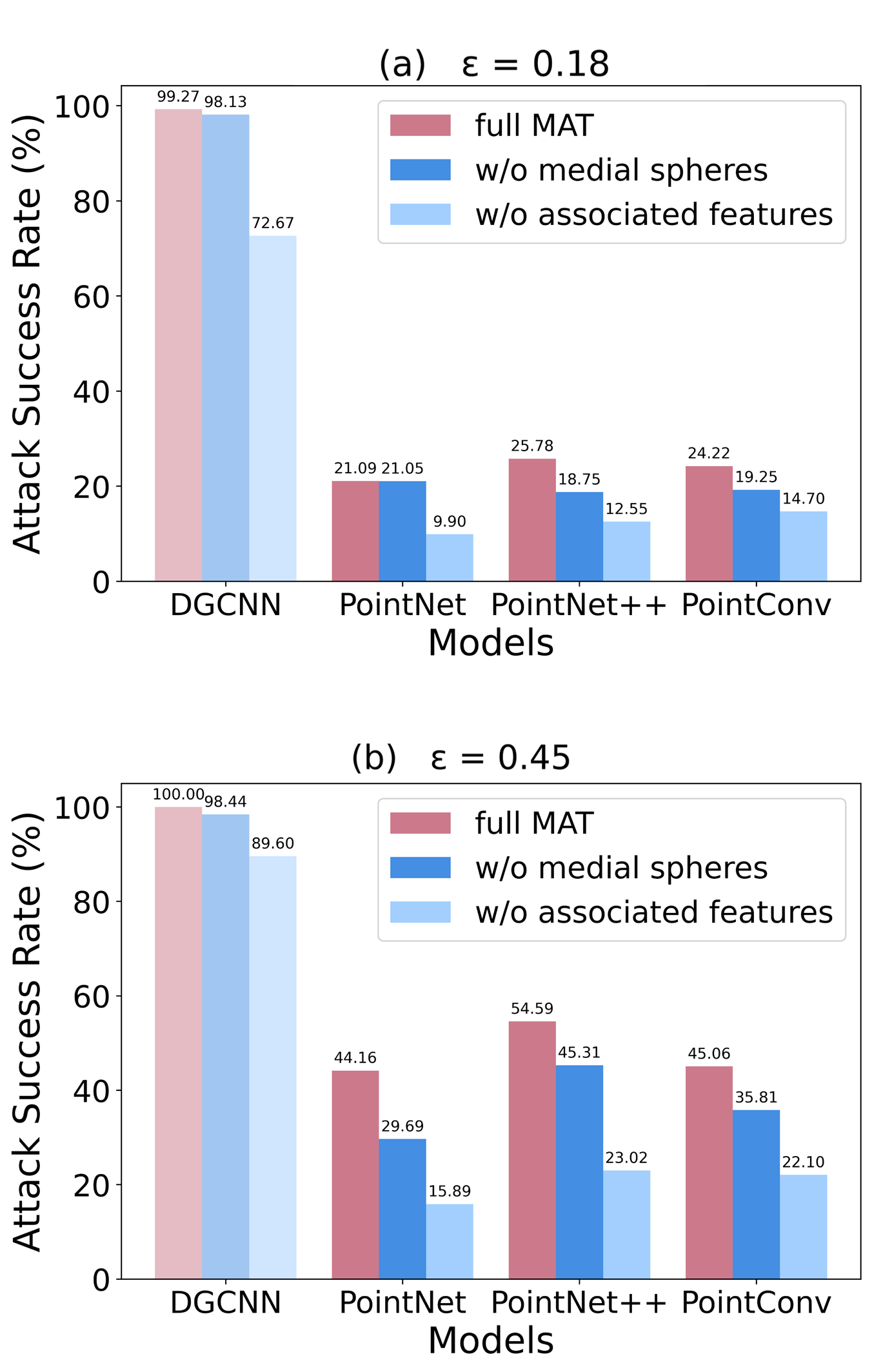}

\caption{
Ablation analysis of attacking different components of the MAT representation, with adversarial examples generated from DGCNN.  
Results include both white-box (DGCNN) and transfer evaluations on ShapeNet Part,  
under $l_{\infty}$-norm perturbation budgets of (a) $\epsilon = 0.18$ and (b) $\epsilon = 0.45$.
}
    \label{fig: ablation}

\end{figure}

\firstpara{Importance of Perturbing $\mathcal{C}, \mathcal{R}$ and $\mathcal{Z}$.}
We conduct ablation studies to evaluate the impact of perturbing different components of the MAT representation. As shown in Fig.~\ref{fig: ablation}, the results indicate that a high transfer attack success rate is achieved only when both the medial spheres, i.e., $\mathcal{C}, \mathcal{R}$, and their associated features, i.e., $\mathcal{Z}$, are perturbed. This underscores the necessity of jointly perturbing these components to enhance the transferability of adversarial attacks in the MAT representation.

\begin{figure*}[!t]
    \centering
    \includegraphics[width=1\linewidth]{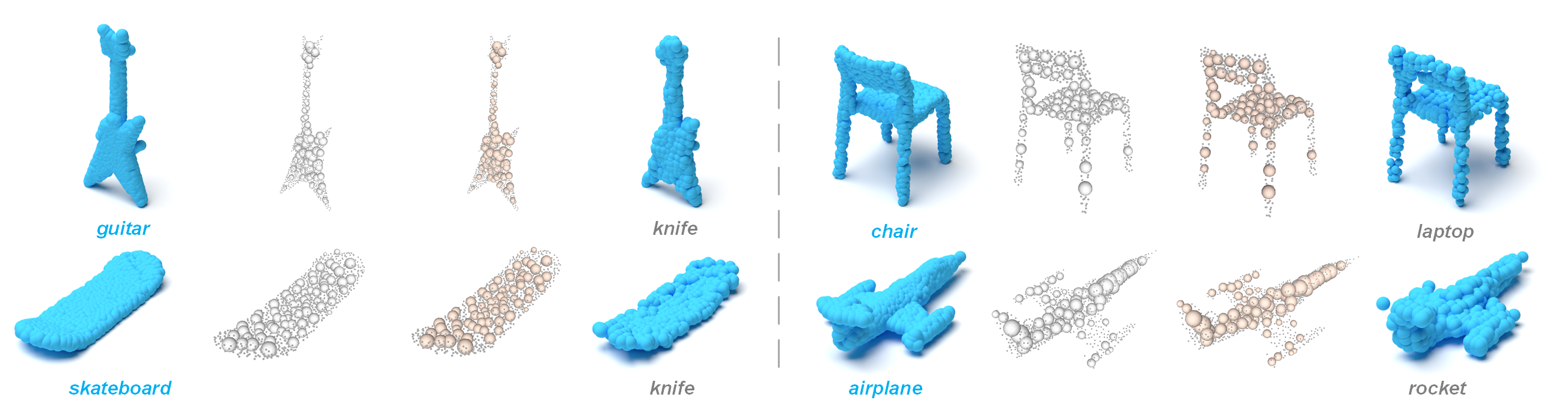}
    \caption{Visualizations of the original point clouds, extracted medial spheres, perturbed medial spheres, and the adversarial point clouds generated by MAT-Adv.  
 The ground truth and predicted labels are marked in blue and gray below the images. }
    \label{fig:medial}
\end{figure*}

\firstpara{Visualization of MAT and Perturbed MAT.}  
To validate the effectiveness of our framework in extracting the medial axis transform (MAT) and utilizing it for generating adversarial point clouds, we present visualizations of the original point clouds, their extracted MAT representations, the perturbed MAT, and the corresponding adversarial point clouds reconstructed from the perturbed MAT, as shown in Fig.~\ref{fig:medial}.
The medial spheres are observed to cover the main body of the object, accurately capturing its global structural characteristics.  
Although the reconstructed adversarial point clouds appear largely similar to the originals in overall shape, the localized perturbations on the MAT introduce subtle structural differences that are sufficient to alter the model’s prediction.  
This visual evidence supports the claim that our perturbation strategy effectively induces inherent adversarialness, while maintaining plausible and interpretable geometric structure.

\begin{figure}[!t]
    \centering
    \includegraphics[width=0.72\linewidth]{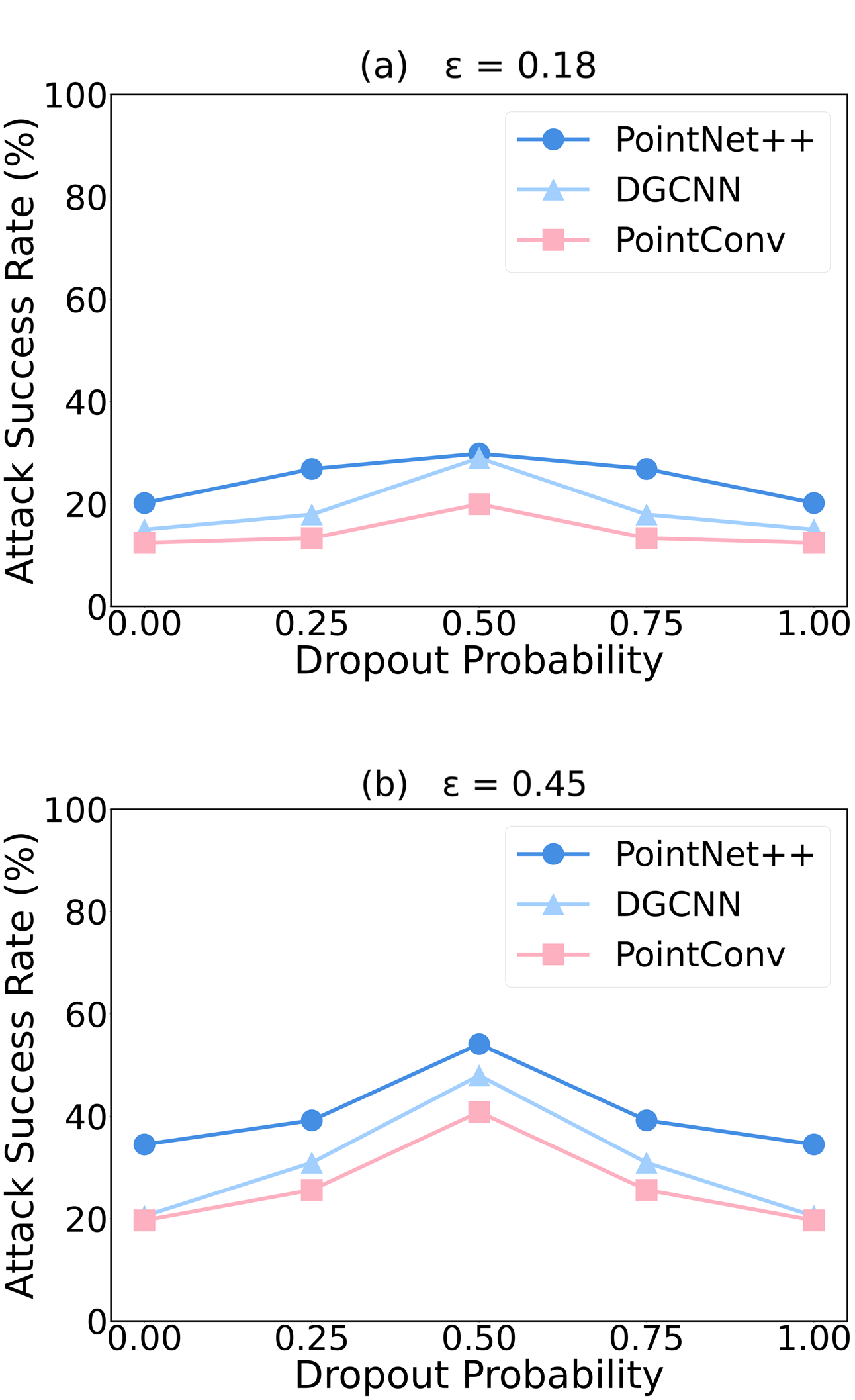}
\caption{
Effect of dropout probability \(\rho\) on transferability, measured by the attack success rate (\%) using adversarial examples generated from PointNet,  
under $l_{\infty}$-norm perturbation budgets of (a) $\epsilon = 0.18$ and (b) $\epsilon = 0.45$.
}
    \label{fig: ablation_roh}
\end{figure}

\firstpara{Influence of Dropout Strategy.}  
We conduct ablation studies to evaluate the effect of the dropout strategy in MAT-Adv. Specifically, we compare variants with and without dropout.  
As shown in Tab.~\ref{tab:ablation-wo-dropout}, removing the dropout mechanism leads to a noticeable decline in transferability, although the overall framework still maintains competitive performance.

We further analyze the impact of the dropout probability \(\rho\).  
As shown in Fig.~\ref{fig: ablation_roh}, we report the attack success rates under different values of \(\rho\), including 0, 0.25, 0.5, 0.75, and 1.  
Based on these results, we select the optimal setting of \(\rho = 0.5\) for our final configuration.

\firstpara{Validation of Intrinsic Properties of MAT Representation.}
To validate the intrinsic nature of the MAT representation, we generate point clouds with varying densities by sampling 4, 6, and 8 points per sphere from a given MAT representation. These generated point clouds are then classified using a classifier trained on the original point cloud.
As shown in Tab.~\ref{tab:skelball accuracies}, the classifier achieves consistently high accuracy across all densities, confirming that point clouds reconstructed from the MAT representation retain intrinsic features of the original data. This result highlights the intrinsic properties embedded in the MAT representation.

\begin{table}[!t]
\centering
\caption{Effect of the dropout strategy on the transferability performance of MAT-Adv from PointNet.}
\label{tab:ablation-wo-dropout}
\setlength{\tabcolsep}{0.5mm}{
\scalebox{0.88}{
\begin{tabular}{@{}c|ccc|ccc@{}}
\toprule
                         & \multicolumn{3}{c|}{$\epsilon=0.18$}             & \multicolumn{3}{c}{$\epsilon=0.45$}              \\ \cmidrule(l){2-7} 
\multirow{-2}{*}{} &  PointNet++ & DGCNN & PointConv                     & PointNet++ & DGCNN & PointConv \\ \midrule
 w/o          &  26.76      & 25.17 & 18.01     & 44.43      & 43.49 & 20.88     \\
w/  & \textbf{29.85} & \textbf{28.98} & \textbf{19.97} &  \textbf{54.11} & \textbf{47.98} & \textbf{40.83} \\ \bottomrule
\end{tabular}}
}
\end{table}

\begin{table}[!t]
\centering
\setlength{\tabcolsep}{3.2mm}{
\caption{
Classification accuracies of various DNN models on the original point clouds from ShapeNet Part and re-sampled point clouds derived from the MAT representation, with 4, 6, and 8 points (pts) sampled per medial sphere (MS).
}
\label{tab:skelball accuracies}
\scalebox{0.88}{
\begin{tabular}{@{}c|cccc@{}}
\toprule
Network   & Original & 4 pts/MS & 6 pts/MS & 8 pts/MS  \\ \midrule
PointNet   & 98.64  & 98.61         & 98.43         & 98.57           \\
PointNet++  & 99.09 & 98.43         & 98.29         & 98.40            \\
DGCNN      & 98.99 & 98.50         & 97.50         & 97.84            \\
PointConv   & 98.82  & 96.93         & 97.60         & 97.63           \\ \bottomrule
\end{tabular}
}
}
\vspace{-3mm}
\end{table}

\firstpara{Imperceptibility.}  
In addition to the two $\ell_{\infty}$-norm perturbation budgets evaluated in the main text, we further assess the imperceptibility of adversarial examples generated with $\epsilon = 0.18$ using several alternative distance metrics, including the Hausdorff distance, the uniform metric~\cite{li2019pu}, and the KNN distance~\cite{tsai-2020-robust(smooth)}.


\begin{table*}[!h]
\centering

\caption{Imperceptibility comparison of different methods on PointNet and DGCNN trained on ShapeNet Part, measured by Hausdorff distance (HD), curvature difference (Curv), and uniformity (Uniform).}
\label{tab:imperceptibility}
\setlength{\tabcolsep}{3.6mm}{
\scalebox{0.98}{
\begin{tabular}{@{}c|cccc|c|cccc@{}}
\toprule
Model & Method    & HD     & Curv   & Uniform & Model & Method    & HD     & Curv   & Uniform \\ \midrule
\multirow{3}{*}{PointNet} & AOF & 0.0893 & 0.1147 & 0.3586 & \multirow{3}{*}{DGCNN} & AOF & 0.0667 & 0.1573 & 0.3889 \\
      & PF-Attack & 0.0508 & 0.0948 & 0.2610  &       & PF-Attack & 0.0481 & 0.1296 & 0.2939  \\
      & Ours      & 0.0553 & 0.0832 & 0.3015  &       & Ours      & 0.0496 & 0.1176 & 0.3214  \\ \bottomrule
\end{tabular}}}
\end{table*}
As shown in Tab.~\ref{tab:imperceptibility}, our method achieves a comparable level of imperceptibility to existing approaches across all three metrics.  
This demonstrates that, while offering improved transferability and robustness, our approach does not compromise on perceptual quality.

\firstpara{Efficiency Analysis.}
To assess efficiency during attack time, we report average runtime per point cloud in Tab.~\ref{tab:running_time}, comparing MAT-Adv with other methods under the same setting. Although we have a slight gap with white-box   attacks, the generation time of our adversarial point cloud is comparable to other black-box methods. 

This confirms that MAT-Adv is computationally practical during deployment and does not incur significant overhead compared to standard gradient-based methods.

\begin{table}[]
\centering
\caption{Time comparison. The source model is PointNet. Running Time is the average time (in minutes) taken to generate one adversarial point cloud.}
\label{tab:running_time}
\scalebox{0.99}{
\begin{tabular}{@{}c|cccc@{}}
\toprule
Attack   Method & 3D-Adv & AOF  & PF-Attack & MAT-Adv \\ \midrule
Running Time    & 0.09   & 0.28 & 0.69      & 0.64    \\ \bottomrule
\end{tabular}}
\end{table}

\section{Conclusion}

In this paper, we have presented MAT-Adv, a novel adversarial attack framework that perturbs the intrinsic medial axis transform (MAT) representation to generate adversarial point clouds with strong transferability and undefendability.  
The rationale behind this design is that manipulating intrinsic geometric representations can induce inherent adversarialness, enabling adversarial examples to remain effective across diverse models and defense mechanisms.  
Extensive experiments demonstrate that MAT-Adv consistently outperforms state-of-the-art methods in both aspects.

\firstpara{Limitation and Future Work.}  
The effectiveness of our approach may be influenced by the quality of the extracted MAT, which can be less stable on shapes with thin structures or complex topology, potentially affecting performance.  
In future work, we will explore more robust intrinsic representations of point cloud geometry to improve the applicability and reliability of adversarial attacks.

\section*{Acknowledgements}
This work was supported in part by the National Natural Science Foundation of China (62472117, 62025207, U2436208), the Guangdong Basic and Applied Basic Research Foundation (2024A1515012064),  
the Science and Technology Projects in Guangzhou (2025A03J0137).




%
%
%
%
%
%
%
%
%
%
%

\bibliographystyle{elsarticle-num}
\bibliography{main}

\end{document}